\documentclass[11pt]{article}
\usepackage{coling2020}
\usepackage{times}
\usepackage{url}
\usepackage{latexsym}

\usepackage{amsmath}
\usepackage{amssymb}
\usepackage{graphicx}
\usepackage{subfig}
\usepackage[ruled,vlined,linesnumbered]{algorithm2e}
\usepackage{multicol}
\usepackage{multirow}
\usepackage{float}
\usepackage{caption}
\usepackage{wrapfig}
\usepackage{hhline}
\usepackage{hyperref}

\colingfinalcopy 

\title{Out-of-Task Training for Dialog State Tracking Models}

\author{Michael Heck, Christian Geishauser, Hsien-Chin Lin, Nurul Lubis, \\\textbf{Marco Moresi, Carel van Niekerk, Milica Ga\v{s}i\'{c}} \\
  Heinrich Heine University Düsseldorf, Germany \\
  \texttt{\{heckmi,geishaus,linh,lubis,moresi,niekerk,gasic\}@hhu.de}}

\date{}

\begin{document}

\maketitle

\begin{abstract}
Dialog state tracking (DST) suffers from severe data sparsity. While many natural language processing (NLP) tasks benefit from transfer learning and multi-task learning, in dialog these methods are limited by the amount of available data and by the specificity of dialog applications. In this work, we successfully utilize non-dialog data from unrelated NLP tasks to train dialog state trackers.  This opens the door to the abundance of unrelated NLP corpora to mitigate the data sparsity issue inherent to DST.
\end{abstract}

\section{Introduction}
\label{sec:introduction}

\blfootnote{
    %
    %
    %
    %
    %
    %
    \hspace{-0.65cm}  
    This work is licensed under a Creative Commons 
    Attribution 4.0 International License.
    License details:
    \url{http://creativecommons.org/licenses/by/4.0/}.
}

The role of the dialog state tracker in a task-oriented dialog system is to summarise the history of the conversation so far and extract the user goal.
Dialog state tracking~(DST) suffers extraordinarily from data sparsity. Collecting data for DST is expensive and time consuming. Typically, conversations are either staged or collected in a Wizard-of-Oz style setup and annotated by hand, severely inhibiting data collection. The enormous number of possible \emph{dialog states} further exacerbates this. 
Even if we combined all data that commercial assistants generate, there will still be realistic but unobserved dialog states.

Unsupervised learning, transfer learning and multi-task learning (MTL)~\cite{caruana1997multitask} in general help mitigate data sparsity. Unsupervised learning relies on predicting inherent characteristics of the data. Transfer learning exploits knowledge learned on related problems to generalize to new tasks. MTL optimizes towards solving multiple tasks at once for synergy effects.   Typically, utilized datasets have related domains and tasks share objectives. 
In other words, these strategies are typically used to address the problem of adaptation.

Recent approaches to adaptation in NLP tasks rely on contextual models. The methods above have been applied to improve generalization across \emph{related} tasks and datasets. For instance,
\newcite{phang2018sentence},~\newcite{wang2019can} and~\newcite{pruksachatkun2020intermediate} facilitate transfer learning by intermediate task fine-tuning (ITFT) on tasks that are related to the target task.
\newcite{peng2020empirical} and \newcite{liu2019multi} jointly optimize transformer based models~\cite{vaswani2017attention} towards multiple related tasks and/or domains. The latter apply MTL to pre-training, rather than fine-tuning.
\newcite{gururangan2020don} report improvements by continuing unsupervised pre-training for domain/task adaptation.
\newcite{raffel2019exploring} and \newcite{keskar2019unifying} propose model architectures that handle diverse tasks with a unified mechanism.

Natural language understanding (NLU) and dialog state tracking (DST) benefit from joint modeling via multitask learning, as this utilizes dialog data more efficiently~\cite{rastogi2018multi}. Recent approaches view DST as a generative problem~\cite{wu2019transferable,ren2019scalable} or as a reading comprehension problem~\cite{gao2019dialog,chao2019bert} utilizing contextual models. In the latter, span prediction or sequence tagging extract relevant information from the input directly. These mechanisms utilize training data more efficiently than early approaches that relied on exhaustive classification given a list of known concepts~\cite{mrkvsic2016neural,liu2017end,zhong2018global}. Pre-training on multiple dialog datasets has been proposed to support subsequent fine-tuning towards specific dialog modeling tasks such as DST~\cite{wu2020tod}. Synergies between DST subtasks can be exploited via MTL~\cite{rastogi2019towards}. Better generalization across slots is attempted via knowledge transfer and zero-shot learning~\cite{rastogi2020schema}. 

All of the above approaches are suitable to better utilize available dialog data, but the general issue of data sparsity persists. There likely will never be enough task-specific data to train dialog models to their full potential. Instead of resorting to the limited quantities of such data, we propose to utilize \emph{non-dialog data} from \emph{unrelated tasks} for the training of DST models.  For this we explore two strategies: (1) in a sequential transfer learning approach, we first train a model to solve an unrelated task, followed by training towards solving DST; (2) we use MTL to jointly optimize towards DST and an unrelated task.  We call our overall approach \emph{out-of-task training} for DST.

With our methods, we achieve new state-of-the-art performance on all four target datasets.
We show that even small amounts of auxiliary task data are beneficial to support model training, especially with MTL, which particularly improves performance on difficult tasks.
Our positive experimental results open the door to the abundance of unrelated NLP corpora defined over a wide range of non-dialog tasks to mitigate the issue of data sparsity in DST.

\begin{figure}[t]
  \centering
  \subfloat[a][Intermediate task fine-tuning (ITFT)]{\includegraphics[page=2, trim=6.5cm 4.5cm 6.5cm 5.3cm, clip=true, width=.47\linewidth,]{Figures.pdf}\label{fig:ft}}%
  \qquad
  \subfloat[b][Multi-task learning (MTL)]{\includegraphics[page=1, trim=6.5cm 4.5cm 6.5cm 5.3cm, clip=true, width=.47\linewidth,]{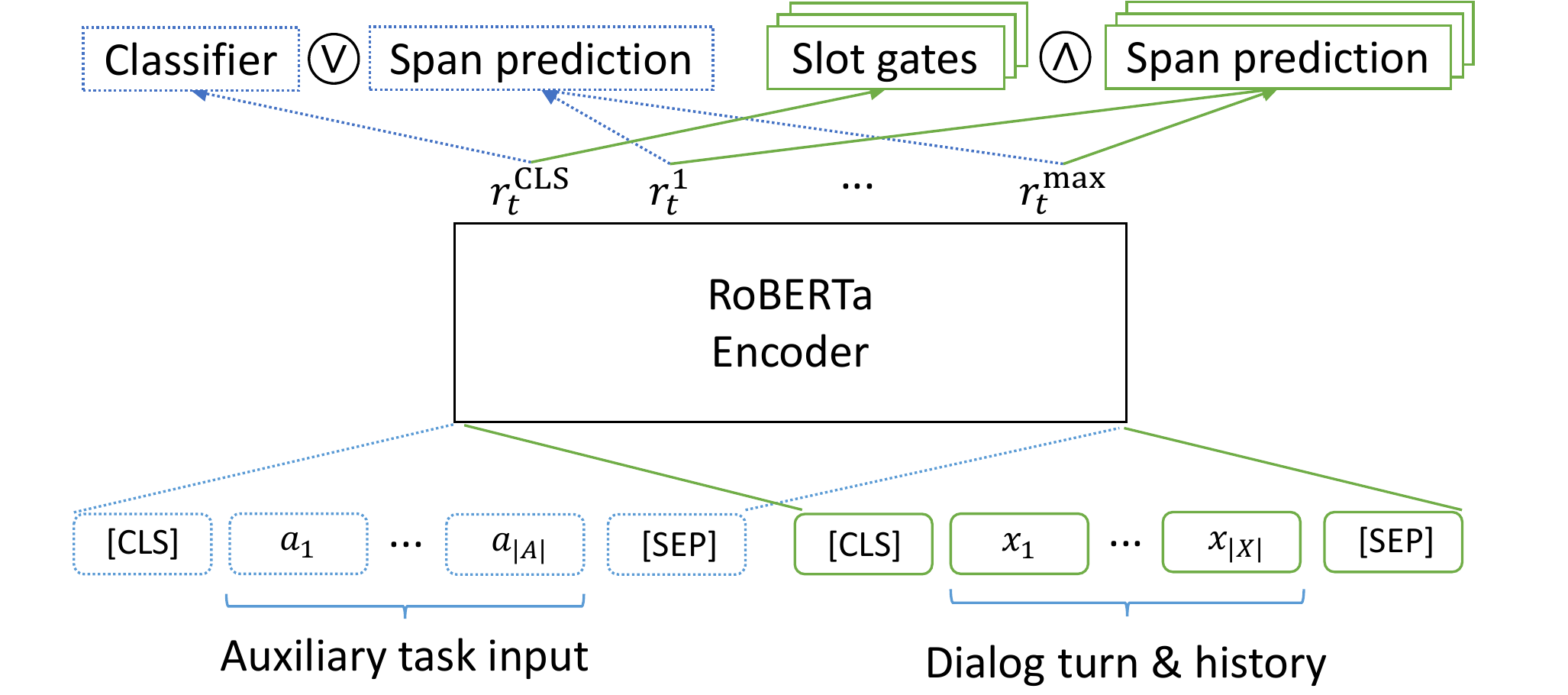}\label{fig:mtl}}
  \vspace{-4pt}
  \caption{Schematics of our proposed out-of-task training schemes. Blue dotted components are used for training on the auxiliary tasks only. Green solid components are used for DST training only.}%
  \label{fig:models}%
  \vspace{-10pt}
\end{figure}

\section{Out-of-Task Training for DST}
\label{sec:method}

\subsection{Dialog State Tracking}
\label{sec:dst}

The task of DST is to extract meaning and intent from the user input, and to keep and update this information over the continuation of a dialog~\cite{young2010hidden}. A restaurant recommender needs to know user preferences such as price, location, etc. These concepts are defined by an ontology in terms of domains (e.g., restaurant), slots (e.g., price range), and values (e.g. expensive).
We utilize TripPy, our publicly available DST model with state-of-the-art performance on a range of datasets.\footnote{Our code is available at \url{https://gitlab.cs.uni-duesseldorf.de/general/dsml/trippy-public}.} The details of this model are described in Heck et al.~\shortcite{heck2020trippy}. We briefly describe the aspects relevant to this work.

TripPy encodes the current dialog context using a transformer model. First, the model determines at each turn whether any of the known domain-slot pairs is present. This is done via slot gates, which either predict that a slot can be filled via a copy mechanism or that it takes a special value (none, dontcare, or true/false). There are three copy mechanisms in TripPy; span prediction and two types of memory lookup. Slot gates and span prediction are realized as classification heads on top of the contextual encoder. The model we use in this work is a modification of the original, as we use RoBERTa~\cite{liu2019roberta} as encoder instead of BERT~\cite{devlin2018bert}. We motivate this by the fact that BERT's distinction of segments has little applicability in dialog. When approaching DST as a reading comprehension task, system and user utterance may take on both the roles of query and response.
The overall performance of this DST model depends on the individual performance of contextual encoder, slot gates and span prediction, i.e., any of these parts could potentially benefit from out-of-task training.

\subsection{Auxiliary Tasks}
\label{sec:tasks}

We consider two types of auxiliary tasks unrelated to DST. The first category encompasses sentence and sentence-pair level classification tasks that aim at discovering linguistic phenomena. We resort to the datasets used by the GLUE benchmark~\cite{wang2018glue}, which cover various NLP problems; (1) MNLI, QNLI, RTE and WNLI for natural language inference (e.g., entailment detection); (2) CoLA for linguistic acceptability classification; (3) SST-2 for sentiment classification; (4) MRPC and QQP for paraphrase detection\footnote{We do not consider STS-B, the semantic textual similarity benchmark due to it being defined as regression problem.}.
We use SQuAD2.0~\cite{rajpurkar2018know}, a question-answering dataset, as representative of token-level classification tasks such as span prediction. SQuaD consists of questions, where the answer to every question is an extractable sequence of text, i.e., a span found in an accompanying paragraph.
We refer to the original papers for further details regarding the datasets. 

We employ the following training constraints: (1) The auxiliary task can either be a classification problem or a span prediction problem, and (2) only one auxiliary task at a time can be used. The latter allows us to clearly identify the effect of particular auxiliary tasks. 

\subsection{Intermediate Task Fine-tuning (ITFT)}
\label{sec:ft}

Our ITFT scheme trains the same model successively on two unrelated tasks, i.e,. the auxiliary task, followed by the DST task. Figure~\ref{fig:ft} is a depiction of the model architecture. The encoder is either followed by task specific classification heads for DST or a task specific classification head for the auxiliary task depending on the training phase.
Both phases of fine-tuning follow the procedure as described by~\newcite{devlin2018bert}.
The intention of ITFT is to steer the encoder's parameters into a favorable direction so that subsequent fine-tuning finds a better local optimum.

\subsection{Multi-task Learning (MTL)}
\label{sec:mtl}

\begin{wrapfigure}{R}{0.34\textwidth}
\vspace{-10pt}
\begin{minipage}{0.34\textwidth}
\begin{algorithm}[H]
\small
\SetAlgoVlined
$m \leftarrow$ pre-trained RoBERTa\\
$B_{\texttt{aux}} \leftarrow$ auxiliary task batches\\
$B_{\texttt{DST}} \leftarrow$ target task batches\\
$e_{\texttt{MTL}} \leftarrow$ last epoch $e$ to do MTL\\
\For{$e\leftarrow 1$ \KwTo $e_{\texttt{max}}$}{
  \For{$s\leftarrow 1$ \KwTo $s_{\texttt{max}}$}{
    \If{$e \leq e_{\texttt{MTL}}$}{
      $b_{\texttt{aux}}^s \leftarrow$ next($B_{\texttt{aux}}$)\\
      $m$.update($b_{\texttt{aux}}^s$)\\
      \If{$b_{\texttt{aux}}^s$ is last}{
        reset($B_{\texttt{aux}}$)\\
      }      
    }
    $m$.update(next($B_{\texttt{DST}}$))\\
  }
  reset($B_{\texttt{DST}}$)\\
}
\caption{MTL}
\label{algo:mtl}
\end{algorithm}
\end{minipage}
\end{wrapfigure}

With MTL, we train the same model simultaneously on two unrelated tasks. Figure~\ref{fig:mtl} is a depiction of the model architecture. The strategy is formally outlined in Algorithm~\ref{algo:mtl}. For each step $s$ that is to be trained on DST, we also train one additional step on the auxiliary task. In other words, the training alternates between auxiliary task and target task on the level of steps. We share one optimizer for both tasks and perform two successive updates (lines 9 and 12 in Algorithm~\ref{algo:mtl}), one for each batch $b$. The number of these double steps is determined by $s_{\texttt{max}}$, the maximum number of steps for the target task, so as to not overpower the main task. A hyperparameter $e_{\texttt{MTL}}$ determines the last epoch for which MTL is applied. If $e_{\texttt{MTL}} < e_{\texttt{max}}$ (maximum number of epochs), then we fine-tune only on DST for the remainder.

\section{Experiments}
\label{sec:experiments}

\begin{table}[t]
  \small
  \centering
  \begin{tabular}{l|c||ll|ll|ll|ll||cc}
   Auxiliary & Available & \multicolumn{2}{c|}{sim-M} & \multicolumn{2}{c|}{sim-R} & \multicolumn{2}{c|}{WOZ 2.0} & \multicolumn{2}{c||}{MultiWOZ 2.1} & \multicolumn{2}{c}{average diff.} \\
   \hhline{~~----------}
   Task & Samples$^1$ & ITFT & MTL & ITFT & MTL & ITFT & MTL & ITFT & MTL & ITFT & MTL \\
   \hline\hline
   - (Baseline) & 2/6/3/57k & 88.8 & 88.8 & 89.1 & 89.1 & 92.1 & 92.1 & 56.2 & 56.2 & - & - \\
   \hline\hline
   SQuaD2.0 & 130k & 91.0$^*$ & \textbf{92.1}$^{**}$ & 89.4 & 90.2$^{**}$ & \textbf{92.9}$^*$ & 92.4 & 56.2 & 56.9$^{**}$ & 0.8 & 1.4 \\
   MRPC  & 3.7k & 89.7 & 90.7$^*$ & 89.8$^*$ & 90.1$^*$ & 92.7$^*$ & \textbf{93.1}$^{**}$ & 55.5 & 57.1$^{**}$ & 0.4 & 1.2 \\
   QNLI  & 108k & 89.6 & 90.9$^*$ & 89.9$^{**}$ & 89.7$^*$ & 92.2 & 92.8$^*$ & \textbf{56.3} & 57.1$^{**}$ & 0.5 & 1.1 \\
   SST-2 & 67k & \textbf{91.2}$^{**}$ & 90.2 & \textbf{90.4}$^{**}$ & 89.7 & 92.2 & 93.0$^*$ & 56.0 & \textbf{57.2}$^{**}$ & 0.9 & 1.0 \\
   QQP   & 364k & 88.5 & 90.5$^*$ & 88.8 & 89.8$^*$ & 91.6 & 93.0$^*$ & 56.0 & 57.0$^{**}$ & -0.3 & 1.0 \\
   CoLA  & 8.5k & 90.9$^{**}$ & 90.1$^*$ & 89.7 & 89.8$^*$ & 92.6 & 92.1 & 56.2 & 57.1$^{**}$ & 0.8 & 0.7 \\
   RTE   & 2.5k & 89.2 & 89.7 & 89.5 & 89.9$^*$ & 92.3 & 92.4 & 55.7 & 57.1$^{**}$ & 0.1 & 0.7 \\
   WNLI  & 634 & 89.5 & 89.4 & 89.9$^{**}$ & 89.4 & 92.4 & 92.3 & 56.2 & \textbf{57.2}$^{**}$ & 0.5 & 0.5 \\
   MNLI  & 393k & 89.2 & 88.3 & 89.2 & \textbf{90.3}$^{**}$ & 92.0 & 92.1 & \textbf{56.3} & \textbf{57.2}$^{**}$ & 0.1 & 0.4 \\
   \hline
   average   & - & 89.9 & 90.2 & 89.6 & 89.9 & 92.3 & 92.6 & 56.1 & 57.1 & 0.4 & 0.9 \\
  \end{tabular}
  \caption{Performance comparison of out-of-task training methods and utilized tasks. Bold indicates best performance per dataset and training method. $^{**}$ and $^{*}$ indicate statistically significant improvements over the baseline with $p<0.05$ and $p<0.1$, respectively. $^1$the maximum use of auxiliary task samples for MTL is $e_{\texttt{max}}$ times the number of samples for the target task (see Section~\ref{sec:mtl}).}
  \label{tab:results}
  \vspace{-13pt}
\end{table}

\paragraph{Datasets} We conduct our evaluation on four dialog datasets. MultiWOZ 2.1~\cite{eric2019multiwoz} is the most challenging, containing over 10k dialogs defined over 5 domains with 30 domain-slot pairs. WOZ 2.0~\cite{wen2016network} is a single-domain benchmark. sim-M and sim-R~\cite{shah2018building} are single-domain datasets that are challenging due to their high out-of-vocabulary (OOV) rate in some slots.

\paragraph{Scoring} We use joint goal accuracy (JGA) on the evaluation sets as the primary measure to compare individual models. JGA is the ratio of dialog turns for which all slots have been filled with the correct value according to the ground truth.
We report average JGA over 5 tests each with different seeds.

\paragraph{Training Details} As encoder we use \emph{RoBERTa-base}.
Training for the first phase of ITFT follows~\newcite{devlin2018bert}.\footnote{For the first phase of SQuaD2.0 ITFT, the maximum input sequence length is 384, the initial LR is $5e^{-5}$, and training is for two epochs. For the first phase of ITFT on GLUE tasks, the initial LR is $2e^{-5}$ and training is for three epochs.}
For target task training and MTL, the maximum input sequence length is 180 tokens after Byte-pair encoding~\cite{sennrich2015neural}. We use Adam optimizer~\cite{kingma2014adam} with joint cross entropy and back-propagate through the entire network. The initial learning rate (LR) is $1e^{-4}$. We conduct training with a warm-up proportion of 10\% and linear LR decay. Weight decay is set to 0.01. During training we use a dropout~\cite{srivastava2014dropout} rate of 30\% on the RoBERTa output, and 10\% on the classification heads. We use early stopping based on the JGA of the development set. $e_{\texttt{max}} = 10$, and $e_{\texttt{MTL}} = 7$. We do not use slot value dropout~\cite{xu2014targeted} except for sim-M.

\subsection{Results}
\label{sec:experiments:ssec:results}

Table~\ref{tab:results} lists our out-of-task training results on all four DST datasets, compared to a baseline that does not use any auxiliary task. It can be seen that additional training on an unrelated task produces considerably better models in almost every tested combination. With one exception, the average JGA of all models trained with IFTF or MTL is always higher than the respective baseline performance.

\paragraph{ITFT vs. MTL}

\begin{figure}[t]
  \centering
  \subfloat[a][ITFT vs. MTL]{\includegraphics[page=5, trim=11.5cm 6cm 9.8cm 6cm, clip=true, width=.47\linewidth,]{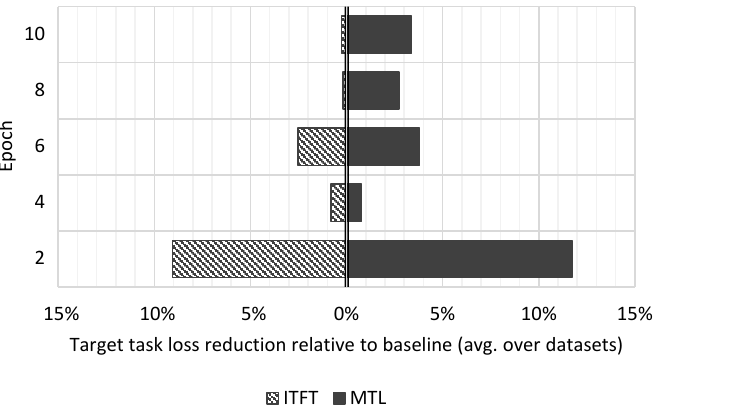}\label{fig:ana1}}
  \qquad
  \subfloat[b][Impact of data amount]{\includegraphics[page=4, trim=10.5cm 6cm 10.8cm 6cm, clip=true, width=.47\linewidth,]{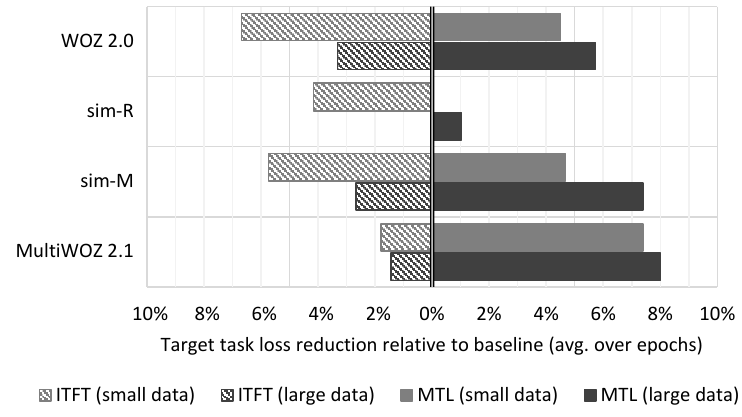}\label{fig:ana2}}%
  \vspace{-4pt}
  \caption{(a) Both methods help early training. In contrast to ITFT, MTL maintains a positive effect on target task training throughout. (b) Loss reductions are averaged over small (WNLI, RTE, MRPC, CoLA; ~4k samples on avg.) and large (SST-2, QNLI, SQuaD2.0, QQP, MNLI; ~212k samples on avg.) out-of-task datasets. MTL benefits from more data, while ITFT performs better with small datasets.}%
  \label{fig:analysis}%
  \vspace{-10pt}
\end{figure}

MTL shows a better performance than ITFT 3 out of 4 times. On every single DST dataset, it is preferrable to use multi-task learning rather than sequential fine-tuning. On average, MTL improves the performance of DST by almost 1\% absolute across all datasets, while ITFT improves the average performance by 0.4\%. In only one case (QQP), ITFT harms performance consistently. In stark contrast, MTL successfully utilizes QQP to consistently improve performance for all DST tasks. Figure~\ref{fig:ana1} shows that both methods benefit early target task training. However, only MTL, which revisits out-of-task data during training, maintains a positive effect throughout all epochs.

\paragraph{Potential impact of target task difficulty}

It is reasonable to assume that target tasks do not benefit equally from out-of-task training, depending on the baseline model's initial capacities. WOZ 2.0 can be considered to be the easiest task to solve. sim-R and sim-M both feature slots with high OOV rates, and sim-M contains extremely limited amounts of data. MultiWOZ 2.1 is most challenging. Table~\ref{tab:results} shows that more difficult tasks tend to benefit more from MTL than easier tasks. This is also true for ITFT except for MultiWOZ. 

\paragraph{Potential impact of data amount}

The amount of target task data and potential improvement via out-of-task training seems uncorrelated. The amount of available and utilized auxiliary task data likewise does not seem to be decisive. Even the smallest of the datasets (WNLI, RTE, MRPC, CoLA), can be utilized successfully to significantly improve DST. However, we did observe a correlation between the auxiliary task data size and the performance of the training methods. Figure~\ref{fig:ana2} shows that MTL tends to benefit from larger out-of-task datasets, while ITFT performs better on small datasets. MTL only sees a subset of all samples of the auxiliary task if the target task is small, yet clearly outperforms ITFT, which always sees all out-of-task training samples and might therefore suffer from adverse effects of over-training on an unrelated task.

\paragraph{Potential impact of auxiliary task type}

Table~\ref{tab:results} is not indicative of trends regarding the usefulness of particular auxiliary task types. We did observe that span prediction (SQuaD), sentiment classification (SST-2) and linguistic acceptability classification (CoLA) led to more consistent improvements than NLI-type tasks and paraphrase detection (MRPC, QQP). The latter two types lead to significantly lower improvements with ITFT, while MTL can benefit from all task types comparably well.

\paragraph{DST training effects}

SQuaD is the only auxiliary task that utilizes the token-level representations of RoBERTa for prediction, while all other tasks (which are GLUE tasks) solely rely on the sequence-level representation. Table~\ref{tab:results_training_effects} shows that fine-tuning on either task category leads to similar DST performance improvements in terms of average slot gate accuracy. While slot gates expect sequence representations as input, span prediction relies on token representations. As can be seen, out-of-task training with SQuaD leads to larger improvements on span prediction than GLUE tasks. The ``movie'' and ``restaurant'' slots in sim-M and sim-R show very high OOV rates (100\% and 40\%). SQuaD proved most helpful to improve accuracies of these particularly difficult slots. Overall, both task categories proved beneficial for improving DST performance.

\begin{table}[t]
  \small
  \centering
  \begin{tabular}{l||ccc|ccc||ccc|ccc}
   & \multicolumn{6}{c||}{ITFT} & \multicolumn{6}{c}{MTL} \\
   \hhline{~------------}
   Auxiliary Task & \multicolumn{3}{c|}{All Slots} & \multicolumn{3}{c||}{High OOV Slots} & \multicolumn{3}{c|}{All Slots} & \multicolumn{3}{c}{High OOV Slots} \\
   \hhline{~------------}
   & SA & SGA & SPA & SA & SGA & SPA & SA & SGA & SPA & SA & SGA & SPA \\
   \hline\hline
   - (Baseline) & 96.5 & 97.6 & 91.5 & 91.4 & 95.3 & 75.7 & 96.5 & 97.6 & 91.5 & 91.4 & 95.3 & 75.7 \\
   \hline\hline
   Avg. All Tasks & 96.8 & 97.6 & 92.2 & 92.0 & 95.5 & 77.9 & 96.9 & 97.6 & 92.2 & 92.2 & 95.4 & 77.8 \\
   \hline
   Avg. GLUE Tasks & 96.8 & 97.6 & 92.2 & 92.0 & 95.6 & 77.8 & 96.8 & 97.6 & 92.1 & 92.0 & 95.4 & 77.4 \\
   SQuaD & 96.9 & 97.6 & 92.5 & 92.5 & 95.3 & 78.6 & 97.3 & 97.8 & 93.1 & 93.3 & 95.8 & 80.4 \\
  \end{tabular}
  \caption{Average slot (SA), slot gate (SGA) and span prediction accuracy (SPA) after ITFT or MTL.}
  \label{tab:results_training_effects}
  \vspace{-13pt}
\end{table}

\section{Conclusion}
\label{sec:conclusion}

We investigated auxiliary out-of-task training for DST and found that model training benefits most from joint optimization, compared to sequential training. Even though auxiliary tasks and target task are domain and task mismatched, our training schemes consistently improve target task performance, regardless of task types or data amounts. We reach state-of-the-art results with considerable improvements on all target datasets. We showed that out-of-task training is suitable to overcome data sparsity issues. In future work we pursue the direction of scaling up to do joint out-of-task training on multiple unrelated auxiliary tasks. We would also like to investigate iterative approaches to out-of-task training, where new data is added to the training during the course of a model's lifetime, rather than training from scratch.

\section*{Acknowledgements}

M. Heck, N. Lubis and C. van Niekerk are supported by funding provided by the Alexander von Humboldt Foundation in the framework of the Sofja Kovalevskaja Award endowed by the Federal Ministry of Education and Research, while C. Geishauser, H-C. Lin and M. Moresi are supported by funds from the European Research Council (ERC) provided under the Horizon 2020 research and innovation programme (Grant agreement No. STG2018\_804636). Computing resources were provided by Google Cloud.

\bibliographystyle{coling}
\bibliography{coling2020}

\newpage

\section*{Appendix A. Out-of-Task Training for DST Using BERT}

Table~\ref{tab:results_bert} summarizes experimental results when using BERT instead of RoBERTa as encoder in TripPy. Even though BERT benefits less from the out-of-task training, the same tendencies as for RoBERTa are observed. One notable difference is the poor performance of SST-2 for BERT. TripPy with BERT uses segment ID 0 for the current user utterance and segment ID 1 for the system utterance plus dialog history, while TripPy for RoBERTa does not distinguish between multiple segments in the input~\cite{devlin2018bert,liu2019roberta}. Out-of-task training on SST-2 might negatively affect TripPy with BERT, because this data consists of single segments instead of pairs. However, the nature of the task (see Table~\ref{tab:tasks}) seems to be relevant, as CoLa - another a single segment classification problem - does not result in such poor performance using BERT. It is noteworthy that the improvements using RoBERTa over BERT for SST-2 are also above average in the official GLUE benchmark leaderboard\footnote{\url{https://gluebenchmark.com/leaderboard}} (while being on average for CoLA), which might indicate a generally higher aptitude of RoBERTa for learning from SST-2.

\begin{table}[h]
  \small
  \centering
  \begin{tabular}{l|c||ll|ll|ll|ll||cc}
   Auxiliary & Available & \multicolumn{2}{c|}{sim-M} & \multicolumn{2}{c|}{sim-R} & \multicolumn{2}{c|}{WOZ 2.0} & \multicolumn{2}{c||}{MultiWOZ 2.1} & \multicolumn{2}{c}{average diff.} \\
   \hhline{~~----------}
   Task & Samples$^1$ & ITFT & MTL & ITFT & MTL & ITFT & MTL & ITFT & MTL & ITFT & MTL \\
   \hline\hline
   - (Baseline) & 2/6/3/57k & 84.4 & 84.4 & 88.1 & 88.1 & 91.6 & 91.6 & \textbf{55.9} & 55.9 & - & - \\
   \hline\hline
   SQuaD2.0 & 130k & 84.7 & \textbf{86.8}$^*$ & 88.2 & \textbf{88.9}$^*$ & 91.2 & \textbf{92.5} & 55.8 & 56.2$^{**}$ & 0.0 & 1.1 \\
   MRPC  & 3.7k & 85.2$^*$ & 86.6$^*$ & 88.5$^*$ & 88.1 & 91.3 & 91.8 & 55.5 & 56.4$^{**}$ & 0.1 & 0.7 \\
   QNLI  & 108k & 84.8 & 85.0 & 88.6 & 88.3 & 91.3 & 92.1 & \textbf{55.9} & 56.6$^{**}$ & 0.2 & 0.5 \\
   SST-2 & 67k & 83.9 & 82.9 & 88.3 & 88.7$^*$ & 91.6 & 91.3 & \textbf{55.9} & 56.0 & -0.1 & -0.3 \\
   QQP   & 364k & 83.6 & 83.7 & 88.8$^*$ & 88.1 & 91.2 & 91.4 & \textbf{55.9} & 56.1 & -0.1 & -0.1 \\
   CoLA  & 8.5k & 85.2$^*$ & 85.2 & \textbf{89.0}$^{**}$ & 88.5 & 91.3 & 91.5 & 55.6 & 56.4$^{**}$ & 0.3 & 0.4 \\
   RTE   & 2.5k & 84.7 & 85.2$^*$ & 88.3 & 88.6 & \textbf{92.0} & 91.6 & 55.7 & \textbf{56.7}$^{**}$ & 0.2 & 0.5 \\
   WNLI  & 634 & 84.0 & 84.3 & 88.1 & 88.4 & 91.3 & 92.2 & 55.3 & 56.3$^*$ & -0.3 & 0.3 \\
   MNLI  & 393k & \textbf{86.2}$^{**}$ & 84.4 & 88.8$^*$ & 87.9 & 91.0 & 91.4 & 55.2 & 56.4$^{**}$ & 0.3 & 0.0 \\
   \hline
   average   & - & 84.7 & 84.9 & 88.5 & 88.4 & 91.4 & 91.8 & 55.6 & 56.3 & 0.1 & 0.4 \\
  \end{tabular}
  \caption{Performance comparison of out-of-task training methods and utilized tasks when using BERT as encoder. Bold indicates best performance per dataset and training method. $^{**}$ and $^{*}$ indicate statistically significant improvements over the baseline with $p<0.05$ and $p<0.1$, respectively. $^1$the maximum use of auxiliary task samples for MTL is $e_{\texttt{max}}$ times the number of samples for the target task (see Section~\ref{sec:mtl}).}
  \label{tab:results_bert}
\end{table}

\section*{Appendix B. Description of Auxiliary Tasks}

\begin{table}[h]
  \small
  \centering
  \begin{tabular}{l|l|l|l|l}
     Task & Type & Cl. & Input & Task description \\
   \hline\hline
    CoLA & Classification & 2 & single & Predicting a sequence's linguistic acceptability for a sequence \\
    SST-2& Classification & 2 & single & Predicting a sequence's positive or negative sentiment \\
    MRPC & Classification & 2 & pair & Predicting semantic equivalence of potential paraphrases \\
    QQP & Classification & 2 & pair & Predicting semantic equivalence of potential paraphrases \\
    MNLI & Classification & 3 & pair & Predicting if a hypothesis for a premise is neutral, contradiction or entailment \\
    QNLI & Classification & 2 & pair & Predicting if a text snippet contains the answer given a question \\
    RTE & Classification & 2 & pair & Predicting if a sentence pair constitutes entailment \\
    WNLI & Classification & 2 & pair & Predicting if statement 2 is true or false, given statement 1 \\
    SQuaD & Span prediction & - & pair & Predicting start and end of answer span given a text and a question \\
   \hline
  \end{tabular}
  \caption{Overview of auxiliary tasks that were utilized for out-of-task training for DST. \emph{Cl.} denotes the number of target classes for a task. \emph{Input} is either a single sequence or a sequence pair.}
  \label{tab:tasks}
\end{table}

\end{document}